\documentclass[conference]{IEEEtran}
\IEEEoverridecommandlockouts
\usepackage{cite}
\usepackage{amsmath,amssymb,amsfonts}
\usepackage{algorithmic}
\usepackage{graphicx}
\usepackage{textcomp}
\usepackage{booktabs}
\usepackage{subcaption}
\usepackage{algorithm}
\usepackage{algorithmic}
\usepackage{float}
\usepackage{multirow}
\usepackage[hidelinks]{hyperref}
\usepackage{xcolor}
\newcommand{\E}{\mathbb{E}}
\def\BibTeX{{\rm B\kern-.05em{\sc i\kern-.025em b}\kern-.08em
    T\kern-.1667em\lower.7ex\hbox{E}\kern-.125emX}}
\begin{document}

\title{Diverse Priors for Deep Reinforcement Learning}
\author{\IEEEauthorblockN{Chenfan Weng}
\IEEEauthorblockA{\textit{Department of Computer Science} \\
\textit{University of College London}\\
London, UK \\
email: ucabc28@ucl.ac.uk}
\and
\IEEEauthorblockN{Zhongguo Li}
\IEEEauthorblockA{\textit{Department of Electrical and Electronic Engineering} \\
\textit{University of Manchester}\\
Manchester, UK \\
email: zhongguo.li@manchester.ac.uk}
}

\maketitle

\begin{abstract}
In Reinforcement Learning (RL), agents aim at maximizing cumulative rewards in a given environment. During the learning process, RL agents face the dilemma of exploitation and exploration: leveraging existing knowledge to acquire rewards or seeking potentially higher ones. Using uncertainty as a guiding principle provides an active and effective approach to solving this dilemma and ensemble-based methods are one of the prominent avenues for quantifying uncertainty. Nevertheless, conventional ensemble-based uncertainty estimation lacks an explicit prior, deviating from Bayesian principles. Besides, this method requires diversity among members to generate less biased uncertainty estimation results. To address the above 
 problems, previous research has incorporated random functions as priors. Building upon these foundational efforts, our work introduces an innovative approach with delicately designed prior NNs, which can incorporate maximal diversity in the initial value functions of RL. Our method has demonstrated superior performance compared with the random prior approaches in solving classic control problems and general exploration tasks, significantly improving sample efficiency.\\
\end{abstract} 


\section{Introduction}
Reinforcement learning (RL) is a sub-field of machine learning that focuses on the decision-making process in an environment, with the goal of maximizing cumulative reward. In RL, agents interact with the environment to gain experience and then learn an acting policy that leads to the highest possible rewards. However, a challenge arises in determining whether an RL agent should prioritize exploring new state/action or exploiting its existing knowledge to maximize rewards, posing a dilemma known as the exploration-exploitation trade-off. Excessive exploration will lead to slower convergence of the learning process and also fail to exploit the knowledge the agent has already acquired, leading to wasted efforts in exploration. Conversely, excessive exploitation will prevent the agent from exploring potentially better actions or states, resulting in suboptimal performance.

Deep Reinforcement Learning (DRL) expands the capabilities of conventional Reinforcement Learning (RL) by leveraging deep neural networks (NNs) to handle high-dimensional inputs, such as raw images. In recent years, DRL has demonstrated remarkable achievements in various domains, such as playing GO \cite{silver2016mastering}, Atari \cite{pmlr-v119-badia20a} and continuous control \cite{lillicrap2015continuous} games. Despite these notable successes in specific domains, its extensive implementation in real-world problems encounters challenges, primarily stemming from sample inefficiency. To solve even simple problems, DRL typically requires millions of interactions with the environment. These interactions are time-consuming and expensive. A reliable solution to tackle this bottleneck is to improve the sample efficiency via exploration, which involves efficiently exploring the environment to acquire informative experiences that expedite the learning process. An intuitive method for exploring more actively and efficiently is to track the uncertainties of $Q$-values. By doing so, the agent can explore highly uncertain state/action spaces while exploiting low-uncertainty ones. From a Bayesian perspective, estimating the uncertainty can be viewed as a form of posterior inference about the optimal value functions. Therefore, the uncertainty-orientated exploration adheres to the only admissible decision rule: Bayesian rule \cite{cox1979theoretical,wald1950statistical}. 

While Bayesian principles offer a key framework for decision-making guidance, calculating the precise posterior distribution is often impractical. The state-of-the-art approaches for quantifying neural network uncertainty are Bayesian Neural Networks (BNNs) \cite{jospin2022hands}, which learn distributions instead of the point estimates over the weights. Nonetheless, adopting Bayesian NNs requires substantial adjustments to the training process and incurs notable computational costs in comparison to standard (non-Bayesian) NNs. A simple and scalable method to estimate the uncertainty is using the ensemble NNs, which has been proven successful in practice \cite{lakshminarayanan2017simple,osband2016deep}. It aggregates the estimates of multiple distinct NNs and the variance of the ensemble’s predictions can be interpreted as its uncertainty. The intuition behind it is intriguing: NNs' predictions converge to the same results (low uncertainty) around data that is frequently observed while the predictions will be diverse (high uncertainty) around the data that is rarely observed. 
        
Although ensemble methods are effective for quantifying uncertainty, they differ from Bayesian principles in the sense that prior knowledge is not explicitly incorporated. This can lead to biased results for uncertainty estimation\cite{osband2018randomized}, especially when the ensemble size is small. In ensemble-based uncertainty estimations, this prior effect can be considered as the diversity source of members. In other words, the initial weights of member NNs should be sampled from the prior distribution of weights before receiving any data. If the prior distribution is concentrated, then all the members in the ensemble will be similar, which will cause underestimations of the uncertainty (overconfidence). Overconfidence is a significant issue in ensemble uncertainty estimation, where members may agree on wrong predictions due to limited diversity. In contrast to this, if the prior distribution can cover a wide range of diverse models, it will lead to less biased uncertainty estimation results, guiding more comprehensive exploration.

While the significance of maintaining diversity in ensemble-based methods is well-recognized (see, e.g., \cite{fortuin2022priors}), limited research effort has been dedicated to developing possible mechanisms to address this problem. We observe that the effect of priors on the agent's initial behavior is closely related to that caused by intrinsic motivation \cite{chentanez2004intrinsically}, which does not depend on the experience or data collected but relies on agent's intrinsic belief in their acting strategy. Typical intrinsic motivation keeps agents' curiosity on novel states and actions to avoid repeatedly exploring known states and actions \cite{bellemare2016unifying}. Akin to the concept of intrinsic motivation, our proposed mechanism for assembling diverse priors can keep agent's curiosity about the diversity of members to avoid acting according to similar opinions from the ensemble in the initial stage. In the later learning stage, the estimated uncertainty will dominate the guidance of exploration and exploitation behaviors.  

Motivated by the above observations, we introduce a novel approach called Bootstrapped DQN with Diversity Prior (BSDP) to improve exploration efficiency in RL. 
BSDP enhances the diversity among ensemble members at the beginning of training. This is a key feature that mitigates the underestimation of uncertainty during the early phases of training, and also acts as an inherent driving force for the agent's initial exploration activities. Unlike other methods, like no prior or random prior designs, BSDP adaptively adjusts its exploration rate according to disagreement among ensemble members, especially in the initial episodes, which consequently achieves high exploratory capability and superior performance. 

\section{Related Work and Contributions}\label{related_work}

    Benefiting from extensive interactions with the environment, DRL has achieved human-level performance in various challenging domains \cite{bellemare2013arcade, silver2016mastering}. In conventional settings of DRL, data samples are deemed as non-expensive, enabling an RL agent to utilize the information from large and comprehensive datasets, whether they are from human-labeled real data or stimulated simulations \cite{ghavamzadeh2015bayesian}. Recently, effective exploration mechanisms have been widely studied to improve the sampling efficiency \cite{ladosz2022exploration, cowen2020samba, lindner2021information}.

    BNNs estimate probability distributions of network parameters, merging the scalability, expressiveness, and predictive performance in NNs \cite{gawlikowski2023survey}. Although BNNs are state-of-the-art methods for quantifying the uncertainty of NNs, the computational cost of BNNs is high and it requires significant modifications in the training process. In contrast, ensemble methods assemble multiple NNs' predictions to provide an estimation of uncertainty. This class of ensemble methods has advantages like scalability, straightforward implementation and computational efficiency \cite{lakshminarayanan2017simple}. 
    
    A crucial aspect of ensemble-based uncertainty estimation involves maximizing the diversity in the behavior of individual members \cite{renda2019comparing}. The former approaches for increasing the diversity include random initialization and data shuffle \cite{lakshminarayanan2017simple}, bagging and boosting \cite{livieris2021ensemble}, data augmentation\cite{nalepa2019training} and ensemble of different network architectures \cite{herron2020ensembles}. However, none of the above approaches utilize prior for diversifying ensemble members like our advocated method BSDP. Our method not only increases the dissimilarity among members but also incorporates diversified priors into the ensemble method to make the uncertainty estimation closer to the concept of Bayesian principles, generating better uncertainty estimation results. 

    The combination of ensemble-based uncertainty estimation and fundamental reinforcement learning algorithms has received growing interest from researchers. Bootstrapped DQN (BS) enhanced DQN with the bootstrapping ensemble methods \cite{osband2016deep}. It has substantially improved cumulative performance across most games in the Arcade Learning Environment and became a basis for many later ensemble-based RL algorithms. Peer et al. \cite{peer2021ensemble} derived the Bootstrapped DQN by increasing the number of DQN in double DQN frameworks\cite{van2016deep}, which improved the mitigation of the overestimation issues. To address the absence of prior in ensemble-based uncertainty estimation and enhance member diversity, Bootstrapped DQN with Random Prior (BSP) added random priors to ensemble members, which demonstrates effectiveness in solving large-scale problems \cite{osband2018randomized}. However, random is insufficient to ensure the diversity of members and sufficient exploration, since random priors for members are sampled from a uniform distribution \cite{osband2018randomized}, which cannot avoid sampling a close prior twice. Our work Bootstrapped DQN with Diversity Prior (BSDP) designed the prior functions to maximize dissimilarity in the initial behaviors of each member, which has significantly increased the sample efficiency in the early stage of training. 
 
\section{Bootstrapped DQN with Diverse Prior (BSDP)}
BSDP uses an ensemble of DQNs to conduct the learning process. To incorporate the priors during training, the estimated $Q$-value for each member is obtained by combining the outputs of a trainable network $f_\theta(x)$ and a fixed prior network $p(x)$,
\begin{equation}\label{eq:BSP}
    Q_{\theta}(x) = \underbrace{f_\theta(x)}_{\text{trainable}} +  \underbrace{p(x)}_{\text{prior}}. 
    \end{equation}
    
By adopting this method, diversity can be introduced into the weights of prior networks while preserving the normal weight initialization procedure (e.g. He initialization \cite{he2015delving}) for trainable networks. With fixed prior, the TD-error $\delta$ for each member in the ensemble to minimize is,
    \begin{equation} \label{eq:TD-error}
   \delta=\biggl(r_t+\gamma\max_{a'\in\mathcal{A}}(f_{\theta^-}+p)(s_t')-(f_\theta+p)(s_t,a_t)\biggr)^2
    \end{equation}
    where $\theta$ is the trainable parameter, $\theta^-$ is the target function's parameter, $p$ is the fixed prior function.

The prior design in BSDP primarily encompasses two aspects: 1) incorporating priors that exhibit maximum dissimilarity in members' softmax outputs when operating in identical states; 2) increasing the squared second derivatives in all the states to make the initial outputs more nonlinear and complex. The intuition behind 1) is to maximize the model disagreement at the beginning of the training to ensure effective exploration among all the members \cite{pathak2019self}. The intuition behind 2) is to increase the nonlinearity and complexity of initial outputs, which is closely related to optimistic initialization \cite{kaelbling1996reinforcement}, i.e., initializing value functions with optimistic values to encourage exploration. Enhancing nonlinearity and complexity can lead to partial states' values not being optimistically initialized by some members. However, when considering the whole ensemble, it is generally possible to ensure that at least some members are optimistically initialized in these states, which can still enhance the exploration. In the whole process of diversity initialization, only prior functions will be optimized, which will remain fixed after agents start to interact with the environment.
   
The pseudocode of Diverse Prior Initialization is demonstrated in Algorithm \ref{alg:DP}. During the implementation, random state data is generated within the state space, and a model from the ensemble is randomly selected. The loss function to minimize is given by, 
\begin{equation}\label{diversity_initalize}
J(p_{j})=\text{KL\_loss}(\epsilon)+\alpha_1\text{NL\_loss}+\alpha_2\text{BD\_loss}
\end{equation}
where $\epsilon$, $\alpha_1$ and $\alpha_2$ are the constants, and the loss functions will be elaborated in the sequel. 

The first term \text{KL\_loss} computes the negated clipped KL divergence between the softmax output distribution of the chosen model and the median softmax output distribution of the entire ensemble,
\begin{equation}\label{kl_loss}
    \text{KL\_loss}(\epsilon)=-\E_s\text{clip}(D_{\text{KL}}(\sigma(Q_j(s))||\sigma(Q(s))),0,\epsilon)
\end{equation}
where $Q_j(s)$ is the $j^{th}$ member's $Q$ output in state $s$ for all actions, $Q(s)$ is the median $Q$ output of the ensemble in state $s$ for all actions, $D_{\text{KL}}(\cdot||\cdot)$ denotes computing the KL divergence\footnote{$D_{\text{KL}}(P \parallel Q) = \sum_{i} P(i) \log\left(\frac{P(i)}{Q(i)}\right)$} between two distributions, $\sigma(\cdot)$ denotes computing the softmax\footnote{$\sigma(x_i) = \frac{e^{x_i}}{\sum_{j=1}^{N} e^{x_j}}$} of the categorical value for each action and $\epsilon$ is the clipping upperbound. The clipping upperbound $\epsilon$ can prevent the priors' output from becoming extremely high, which may slow down and destabilize RL training.

The second term \text{NL\_loss} computes the negated squared second derivatives,
\begin{equation}\label{non_linear}
    \text{NL\_loss}=-\E_s|Q''_j(s)|^2
\end{equation}
where $|\cdot|$ denotes compute the modulus of the vector and $Q''_j(s)$ is the second derivative of $Q$-value on state $s$ for all the actions, which is approximated by finite-difference methods\footnote{For example, the second derivative of $f(x)$ can be approximated by $\frac{f(x+2h)-2f(x+h)+f(x)}{h^2}$} (FDM).

The third term \text{BD\_loss} computes the squared output of $Q$ for all members,
\begin{equation}\label{bounding_loss}
    \text{BD\_loss}=\E_s|Q(s)^2|
\end{equation}
This term will penalize the large absolute values of $Q$ outputs, which may potentially make the training slow and unstable. Compared to the clipping in the first term, the third term directly constrains the output. 

\begin{algorithm}
\caption{Diverse Prior Initialization}
\label{alg:DP}
\begin{algorithmic}[1]
\STATE \textbf{Initialize} an ensemble of priors with $K$ members $\{p_k\}^{K}_{k=1}$; an ensemble of trainable functions with $K$ members  $\{f_k\}^{K}_{k=1}$
    \FOR{$i=0,1,2,\ldots,M$}
        \STATE sample a state $s_i$ from the state space $\mathcal{S}$: $s_i\sim \mathcal{S}$ 
        \STATE randomly select one model index $j$ from $1,2,\ldots,K$
        \STATE compute $Q$-value for sampled state using selected model: $Q_j(s_i)=f_j(s_i)+p_j(s_i)$
        \STATE compute the median $Q$-value of the ensemble: $Q(s_i)=\text{median}(f(s_i)+p(s_i))$
        \STATE use gradient descent to update $p_j$ by minimizing loss function: $J(p_j)$  
    \ENDFOR
\end{algorithmic}
\end{algorithm}
The pseudocode of BSDP is shown in Algorithm \ref{alg:bootstrapped-dqn}. The Diverse Prior Initialization is completed in the initialization stage of BSDP and the priors will remain unchanged in the later training. For every episode, the agent selects a model $Q_k$ from the ensemble $\{Q_k\}^{K}_{k=1}$ to perform actions and gather data. Each member's prediction on $Q$-value can be considered as the sample from the posterior of the $Q$-value. This concept bears a significant connection to Thompson sampling \cite{russo2018tutorial}, which selects actions based on their estimated probability of being the best choice. The collected data is then shared partially among the ensemble members through a data buffer $B$ with the mask $m_t$, and each member observes different subsets of the overall data, known as bootstrapping \cite{efron1982jackknife}. The mask $m_t$ determines whether each member value function $Q_k$ should be trained using the experience generated at step $t$. In its simplest form, $m_t$ is a binary vector of length $K$ that samples from a $K$-dimensional Bernoulli distribution. For example, $m_t=(1,0,1,1,1)$ means that only the second $Q$-network cannot train on step $t$'s data. It is worth noting that if all the $m_t$ is filled with $1$, then the algorithm degrades to an ensemble method that all the members share the same global dataset. The update process for the $Q$-network is similar to conventional DQN, i.e., minimizing the TD-error defined in \eqref{eq:TD-error}, with the exception of sampling from the masked buffer described above. 
\begin{algorithm}
\caption{Bootstrapped DQN with Diverse Prior}
\label{alg:bootstrapped-dqn}
\begin{algorithmic}[1]
\STATE \textbf{Initialize} an ensemble of $Q$-functions with $K$ members $\{Q_k\}^{K}_{k=1}$ using Diverse Prior Initialization (Algorithm \ref{alg:DP}); Masking distribution $M$; Replay buffer $B$.
\REPEAT
    \STATE reset environment
    \STATE receive initial state $s_0$
        \STATE pick a $Q$-value function to follow $k\sim \text{uniform}\{1,\ldots,K\}$
    \WHILE{episode not terminated}
        \STATE take action $a_t=\arg\max_aQ_k(s_t,a)$
        \STATE receive state $s_{t+1}$ and reward $r_t$ from environment 
        \STATE sample bootstrap mask $m_t \sim M$
        \STATE add $(s_t, a_t, r_{t}, s_{t+1}, m_t)$ to replay buffer $B$
        \STATE sample $K$ mini-batch according to masks from $B$ to update each member $Q$-function by minimizing \eqref{eq:TD-error}.
    \ENDWHILE
\UNTIL{convergence}
\end{algorithmic}
\end{algorithm}

\subsection{Impact of Diverse Prior Initialization}
To illustrate the impact of diverse prior initialization and the influence of each term in the loss function \eqref{diversity_initalize}, we have set up a simple environment. This environment features a one-dimensional state space ranging from $-5$ to $5$ and a binary action space. There are $10$ $Q$-functions in our ensemble, each comprising two neural networks: one trainable and one prior. We will analyze the impact of varying priors on the distribution of the member $Q$-functions. It should be noted that only $Q$-values for action $0$ will be plotted since the $Q$-values for other actions exhibit similar distribution properties.

\begin{figure}[t] 
    \centering
  \subfloat[random prior\label{fig:a}]{%
       \includegraphics[width=0.45\linewidth]{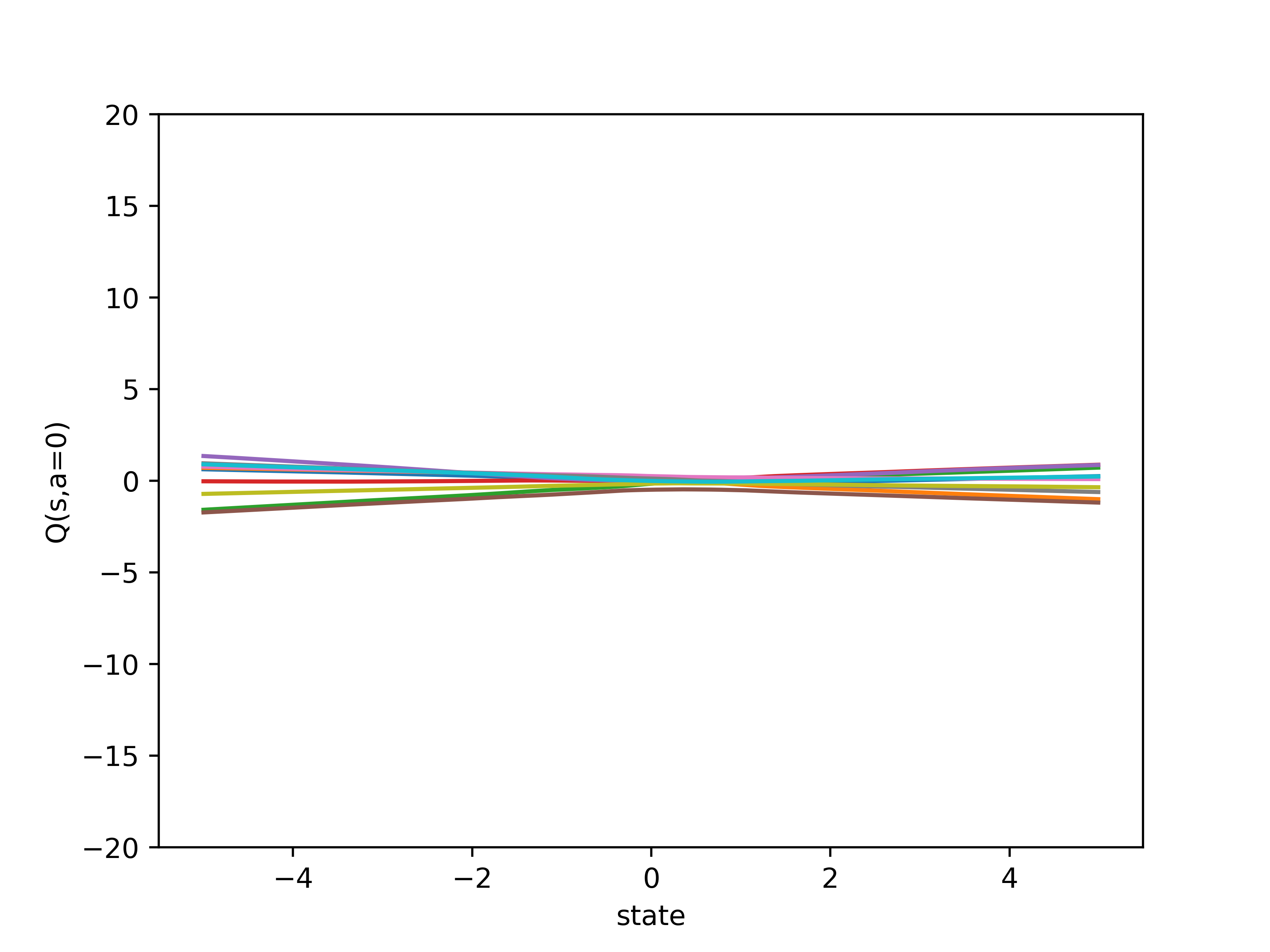}}
  \subfloat[KL\_loss\label{fig:b}]{%
        \includegraphics[width=0.45\linewidth]{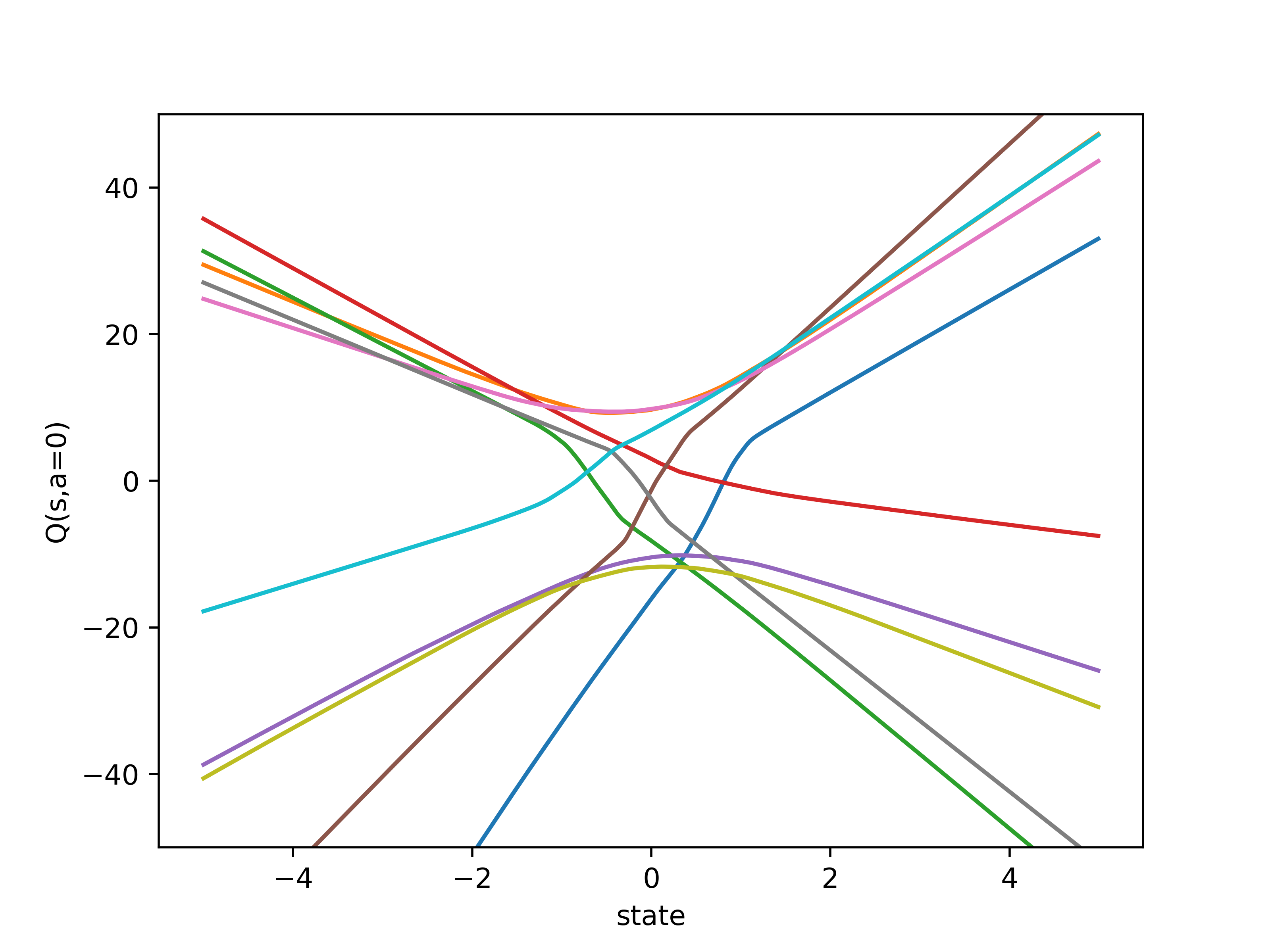}}
    \\
  \subfloat[KL\_loss+BD\_loss\label{fig:c}]{%
        \includegraphics[width=0.45\linewidth]{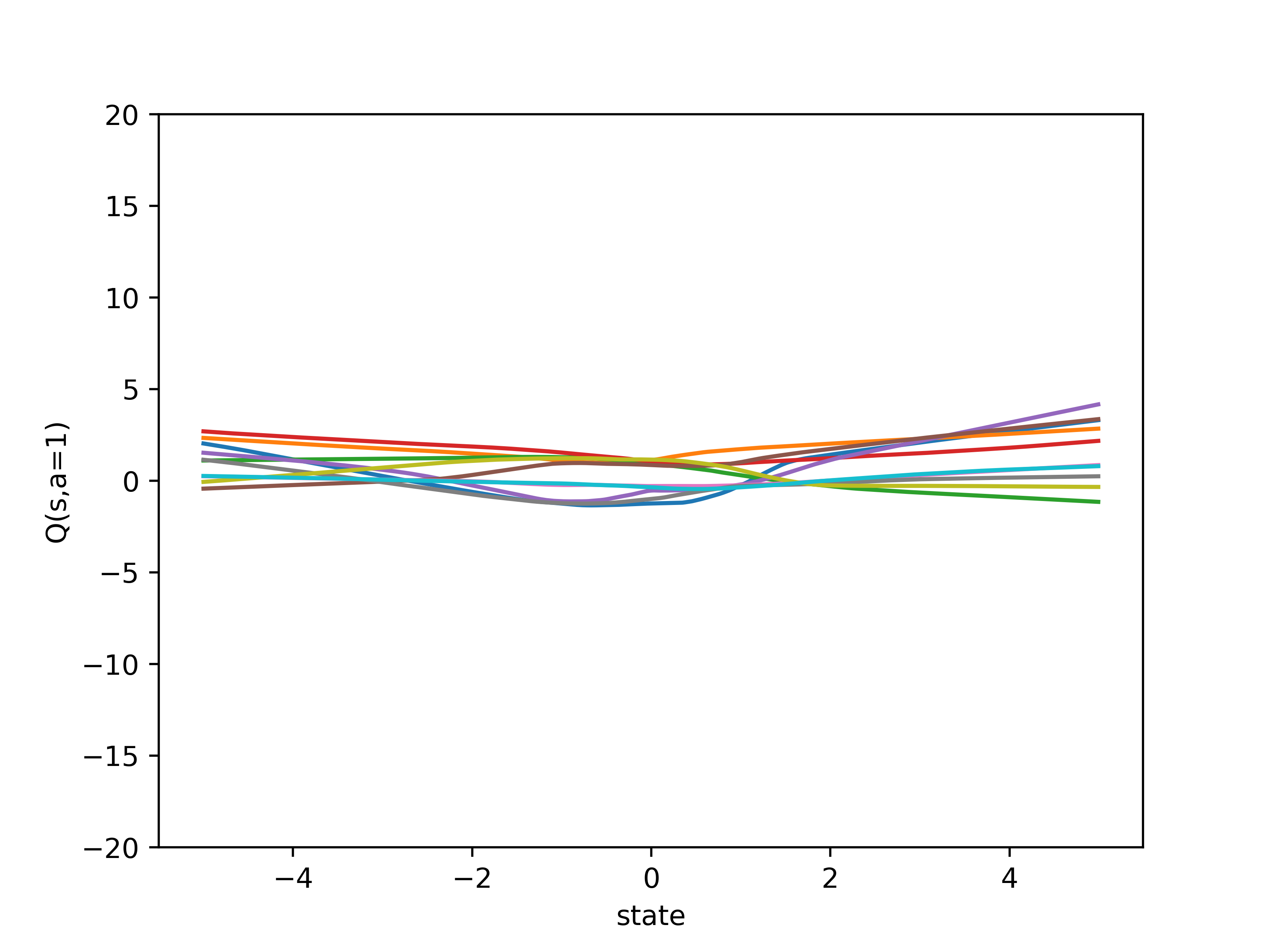}}
  \subfloat[all the three losses \label{fig:d}]{%
        \includegraphics[width=0.45\linewidth]{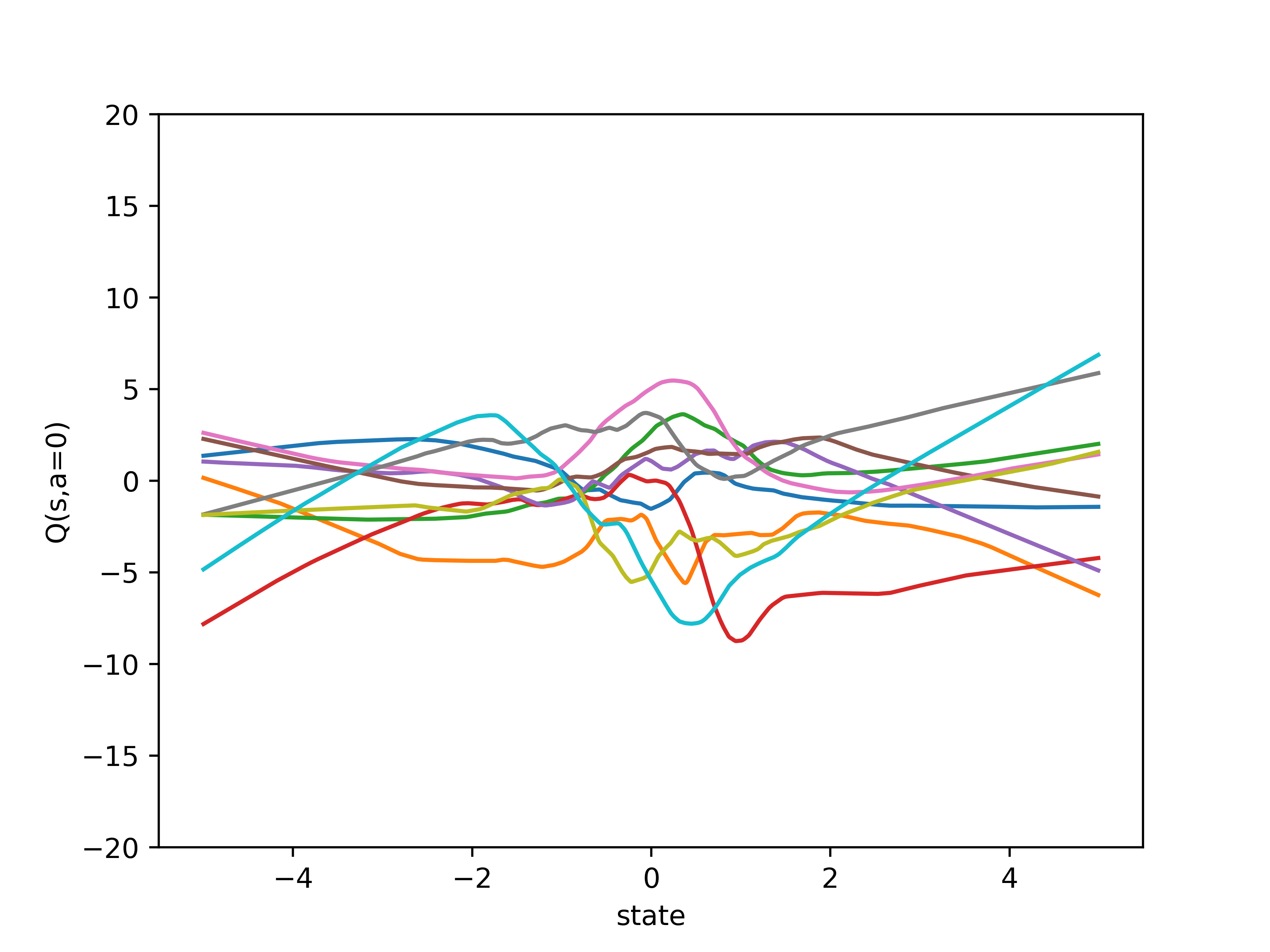}}
         \caption{Random prior outputs v.s. diversity prior outputs: (a) shows the initial $Q$-values for 10 ensemble members using random prior. (b), (c) and (d) show the initial $Q$-values for 10 ensemble members using diverse prior trained on KL\_loss, KL\_loss+BD\_loss and KL\_loss+NL\_loss+BS\_loss respectively.}
         \label{fig:BSP_BSDP_prior}
\end{figure}

The ensemble with random prior is constructed as follows, the total $20$ NNs are initialized randomly using He initialization \cite{he2015delving} which samples the weights from the centralized Gaussian distribution with standard deviation $\sqrt{2/n}$, where $n$ is the input feature number. The $Q$-values for action $0$ across the state space (from $-5$ to $5$) are illustrated in Fig.~\ref{fig:a}, where the curves in different colors represent different members' $Q$-values. Random priors bring limited differences in estimating $Q$-values for members, but $Q$-functions tend to be linear in the state space, with most curves centered at $Q=0$ at state $0$.

First, the diverse prior initialization with only KL\_loss ($\epsilon=0.1$, $\alpha_1=0$, $\alpha_2=0$) is applied to this ensemble, of which the $Q$-functions for members are shown in Fig.~\ref{fig:b}. The $Q$-functions in the ensemble exclude each other and reach large absolute values. However, the output curve for each member is almost linear like Fig.~\ref{fig:a}. After adding $\text{BD\_loss}$ ($\epsilon=0.1$, $\alpha_1=0$, $\alpha_2=0.1$), the result is demonstrated in Fig.~\ref{fig:c}. It is observed that the generated result seems a squeezed version of Fig.~\ref{fig:b} towards zero. Subsequently, the $\text{NL\_loss}$ is added ($\epsilon=0.1$, $\alpha_1=1$, $\alpha_2=0.1$), which increases the complexity of the $Q$-function for each member while also preserving a high dissimilarity among members, as shown in Fig.~\ref{fig:d}. In Section \ref{result_and_discussion}, experimental results will show that this additional diversity introduced by diverse prior initialization can boost the learning process and improve the effective exploration of RL problems.

\section{Results and Discussion}\label{result_and_discussion}
   \begin{table*}
    \centering
    \caption{Experiment details for investigating prior effect}
    \label{tab:rl-training-details}
    \scalebox{0.8}{
    \begin{tabular}{cccccccc}
        \toprule
        Algorithm & Episode & Repeat & Env & Learning Rate &Ensemble size& Diversity Prior & 
        Decaying $\epsilon$-greedy\\
        \midrule
        \multirow{3}{*}{\textbf{BS}}
        & 500 & 5  & Mountain Car& $0.0001$& 5& \textbackslash & \textbackslash\\
        & 500 & 5 & Acrobot & $0.0001$& 5& \textbackslash & \textbackslash\\
        & 500 & 5  & CartPole & $0.0001$& 5& \textbackslash & \textbackslash\\
        & MAX 5000  & 5  & BinaryChain $(n=1\sim20)$ & $0.05$& 10& \textbackslash & \textbackslash\\
        \midrule
        \multirow{3}{*}{\textbf{BSP}}
        & 500 & 5  & Mountain Car& $0.0001$& 5& \textbackslash  & \textbackslash \\
        & 500 & 5  & Acrobot & $0.0001$& 5& \textbackslash  & \textbackslash\\
        & 500 & 5  & CartPole& $0.0001$& 5& \textbackslash & \textbackslash\\
        & MAX 5000  & 5  & BinaryChain $(n=1\sim20)$& $0.05$& 10& \textbackslash & \textbackslash\\
        \midrule
        \multirow{3}{*}{\textbf{BSDP}} 
        & 500 & 5  & Mountain Car& $0.0001$& 5& $\epsilon=0.1$, $\alpha_1=1$, $\alpha_2=0.1$ & \textbackslash  \\
        & 500 & 5  & Acrobot& $0.0001$& 5& $\epsilon=0.1$, $\alpha_1=1$, $\alpha_2=0.1$  & \textbackslash \\
        & 500 & 5  & CartPole& $0.0001$& 5& $\epsilon=0.1$, $\alpha_1=1$, $\alpha_2=0.1$  & \textbackslash \\
        & MAX 5000 & 5  & BinaryChain $(n=1\sim20)$& $0.05$& 10& $\epsilon=0.1$, $\alpha_1=1$, $\alpha_2=0.1$  & \textbackslash\\
        \midrule
        \multirow{3}{*}{\textbf{$\epsilon$-greedy DQN}} 
        & 500 & 5  & Mountain Car& $0.0001$& \textbackslash& \textbackslash & $\beta_1=0.05$, $\beta_2=0.9$, $\lambda=1000$ \\
        & 500 & 5  & Acrobot& $0.0001$& \textbackslash& \textbackslash  & $\beta_1=0.05$, $\beta_2=0.9$, $\lambda=1000$ \\
        & 500 & 5  & CartPole& $0.0001$& \textbackslash& \textbackslash  & $\beta_1=0.05$, $\beta_2=0.9$, $\lambda=1000$ \\
        & MAX 5000 & 5  & BinaryChain $(n=1\sim20)$& $0.05$& \textbackslash& \textbackslash & $\beta_1=0.05$, $\beta_2=0.9$, $\lambda=1000$\\
        \midrule
        \multirow{1}{*}{\textbf{Random}} 
        & MAX 5000 & 5  & BinaryChain $(n=1\sim20)$& \textbackslash & \textbackslash & \textbackslash & \textbackslash\\
        \bottomrule      
    \end{tabular}}

\end{table*}
    To demonstrate the effect of diverse prior, we have conducted an experiment on four RL problems using different approaches: Random\footnote{Random approach means that the agent uniformly samples an action from the action space to conduct for each step}, $\epsilon$-greedy DQN\footnote{The probability for taking exploratory action is $\epsilon=\beta_1+(\beta_2-\beta_1)e^{(k/\lambda)}$}, Bootstrapped DQN (BS), Bootstrapped DQN with Random Prior (BSP) and Bootstrapped DQN with Diverse Prior (BSDP). These four RL problems include  BinaryChain, Mountain Car, Acrobot and CartPole, where the first problem is delicately designed to test the exploration capability of different algorithms and the last three are classic control problems used to compare the overall performance. The training and hyperparameters details are shown in Table \ref{tab:rl-training-details}. ``Repeat" means the same learning will be repeated for a certain number to avoid the influence of occasional results. BinaryChain consists of 20 environments that vary the size from 1 to 20, and the maximum allowed episode number to solve the BinaryChain problem is 5000. Also, to improve the exploration efficiency in BinaryChain problems, for BS, BSP and BSDP, we sample a random member from the ensemble to conduct action for each step instead of each episode like conventional ones.
    
     To be noticed, the curves in the results were averaged over 5 random seeds and the shadow areas were bounded by $\pm\frac{3}{10}$ standard deviations. Besides, to demonstrate the main trend of curves, all the curves for \hyperref[classiccontrolenv]{classic control environments} were smoothed by moving average with a window size of $50$.

    \subsection{BinaryChain Problems}
    In RL, a chain-like environment refers to an environment that can be represented as a chain of states and actions. In such an environment, an agent moves through a series of states and takes specific action for the state transition. This kind of environment can be used to highlight the need for deep exploration by placing the only reward at the end of the chain. With the increase in the length of the chain, it will be exponentially difficult for an agent with a random exploration policy to solve since any mistakes will lead to failure. 
    
\begin{figure}[t]
        \centering
        \includegraphics[width=0.8\linewidth]{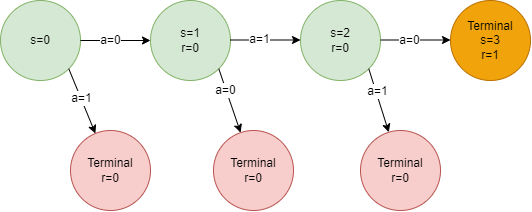}
        \caption{BinaryChain example $N=3$.}
        \label{BinaryChain example}
    \end{figure}

    Our chain-like environment is called BinaryChain and its difficulty is indicated by the size of the chain $N$. Before each game, the environment will generate a single ground truth action trajectory for all the timestep, which is an $N$-length binary vector (e.g. $<0,1,1,0,1,\ldots>$). The agent will be placed at state $s=0$ at the beginning of the episode. For each time step, the agent can choose the action from $0$ and $1$. If the selected action met the ground truth action at this timestep, then the agent can be transited to the next state $(s'=s+1)$. Otherwise, the episode will be terminated with a reward of $0$. The agent will only be rewarded $1$ for acting exactly as the ground truth trajectory, which can lead the agent to state $s=N$. For illustration, a BinaryChain environment of $N=3$ with a ground truth action sequence $<0,1,0>$ is shown in Fig.~\ref{BinaryChain example}.
    
    Theoretically, algorithms without deep exploration (e.g. random exploration) take $\Omega(2^N)$ episodes to find the first reward. Therefore, this environment can be straightforward to test the deep exploration performance of RL methods by varying the size $N$. For BinaryChain problems, episode-number-to-solve was used as the metric, which indicates how many episodes a method requires to reach the positive reward.

\begin{figure}[t]
    \centering    \includegraphics[width=0.9\linewidth]{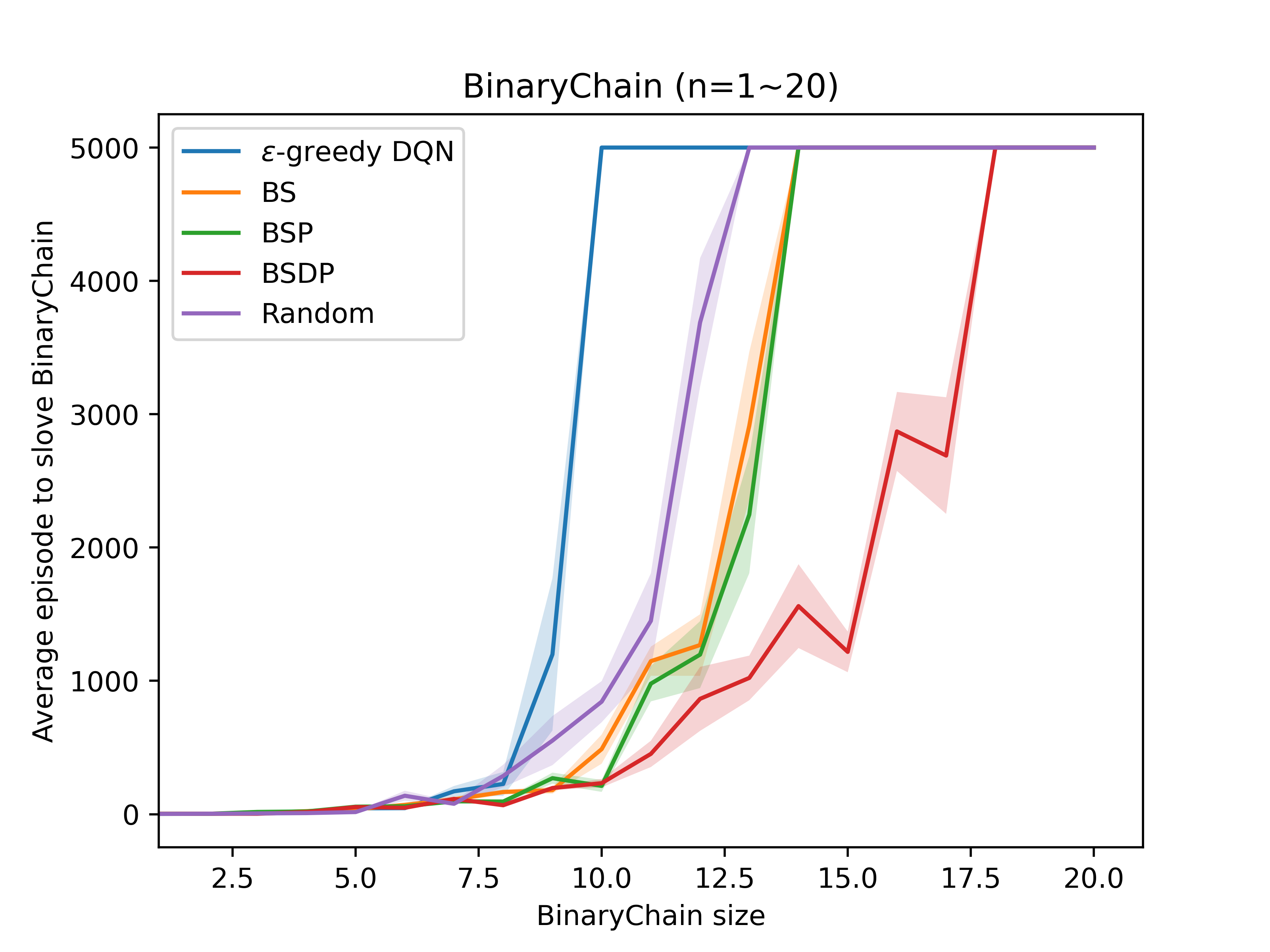}
    \caption{Results for prior effect: BinaryChain (episode-number-to-solve).}
    \label{fig:binary_prior}
\end{figure}

    Fig.~\ref{fig:binary_prior} shows the episode-number-to-solve curves in the \hyperref[binarychain]{BinaryChain} environment for 5 algorithms. We can observe that the pure random policy is better than the $\epsilon$-greedy exploration strategy in this problem, as the former solves BinaryChain-$11$ while $\epsilon$-greedy can only solve BinayChain-$8$ within $5000$ episodes. This is because the exploration behavior in $\epsilon$-greedy is the same as the random policy but it has a lower rate for exploration than pure random policy. Moreover, the updating of the $Q$-value using the Bellman equation cannot provide any exploration advantage in DQN. This indicates that in a sparse reward environment, before finding the first positive reward, a higher exploration rate is more beneficial. 
    
    Benefiting from the uncertainty, the ensemble-based method is more efficient than pure random policy. Among all the ensemble-based methods, the number of required episodes for solving BinaryChain using BS and BSP is similar, while BSDP outperforms both of them significantly. This highlights the importance of diversity prior in enabling deep exploration of the environment with spare rewards.  As shown in Fig.~\ref{fig:binary_prior}, BSDP solves BinaryChain-$17$ within 5000 episodes, which manifests that BSDP can solve more complex sparse reward exploration tasks than BS and BSP, given the exponential increase in the space of decision chains.

\subsection{Classic Control Problems}
    This set of environments is used to rigorously evaluate algorithms' capabilities across classic control problems, encompassing both sparse reward (Mountain Car and Acrobot) and dense reward (Cart Pole) scenarios. Brief descriptions of the environments are listed below. For more detailed information, please refer to the OpenAI Gym's official website \cite{gymlibrary}. For classic control problems, episodic rewards are employed as the metrics to evaluate performance. The exploration rates (E-rate) are also used to demonstrate the ratio of explorative actions, which can indicate agents' exploration behavior and uncertainty about the environment.
    \begin{itemize}
        \item \textbf{Mountain Car}\\
        The Mountain Car MDP is a deterministic environment where a car is randomly positioned at the base of a sinusoidal valley. The goal of the MDP is to strategically accelerate the car, aiming to reach the target state located atop the right hill. The agent will be penalized a $-1$ reward for each step it takes until it reaches the goal. The agent can observe the position (along the x-axis) and velocity. It can choose from three actions for each step: acceleration to left, no acceleration and acceleration to right.  
        \item \textbf{Acrobot}\\
        The system consists of two links connected linearly to form a chain, with one end of the chain fixed. The joint between the two links is actuated. The goal is to apply torques on the actuated joint to swing the free end of the linear chain above a given height from the initial state of hanging downwards. The observation space includes 6 elements: $\cos\theta_1$, $\sin\theta_1$, $\cos\theta_2$, $\sin\theta_2$, $\dot{\theta}_1$ (angular velocity of the joint $\theta_1$) and $\dot{\theta}_2$ (angular velocity of the joint $\theta_2$), where $\theta_1$ is the angle of first joint and $\theta_2$ is angle of the second joint relative to the first link. The action space consists of three elements: applying $-1$, $0$ and $1$ torque to the actuated joint. The agent will receive a $-1$ reward for each step it takes until the free end reaches the goal height.
        \item \textbf{Cart Pole}\\
        A pole is attached by an unactuated joint to a cart, which moves along a frictionless track. The pole is initially positioned in an upright position on the cart, and the objective is to maintain balance by exerting forces in left or right directions on the cart. For each step that the pole is held upright, the agent will receive a reward of $1$ until the pole angle or cart position reaches the limitation. The observation space consists of cart position, cart velocity, pole angle and pole angular velocity. There are only two actions: push the cart to the left or right.
    \end{itemize}
    
\begin{figure} 
    \centering
  \subfloat[Acrobot: reward\label{fig:res1:subfigA}]{%
       \includegraphics[width=0.5\linewidth]{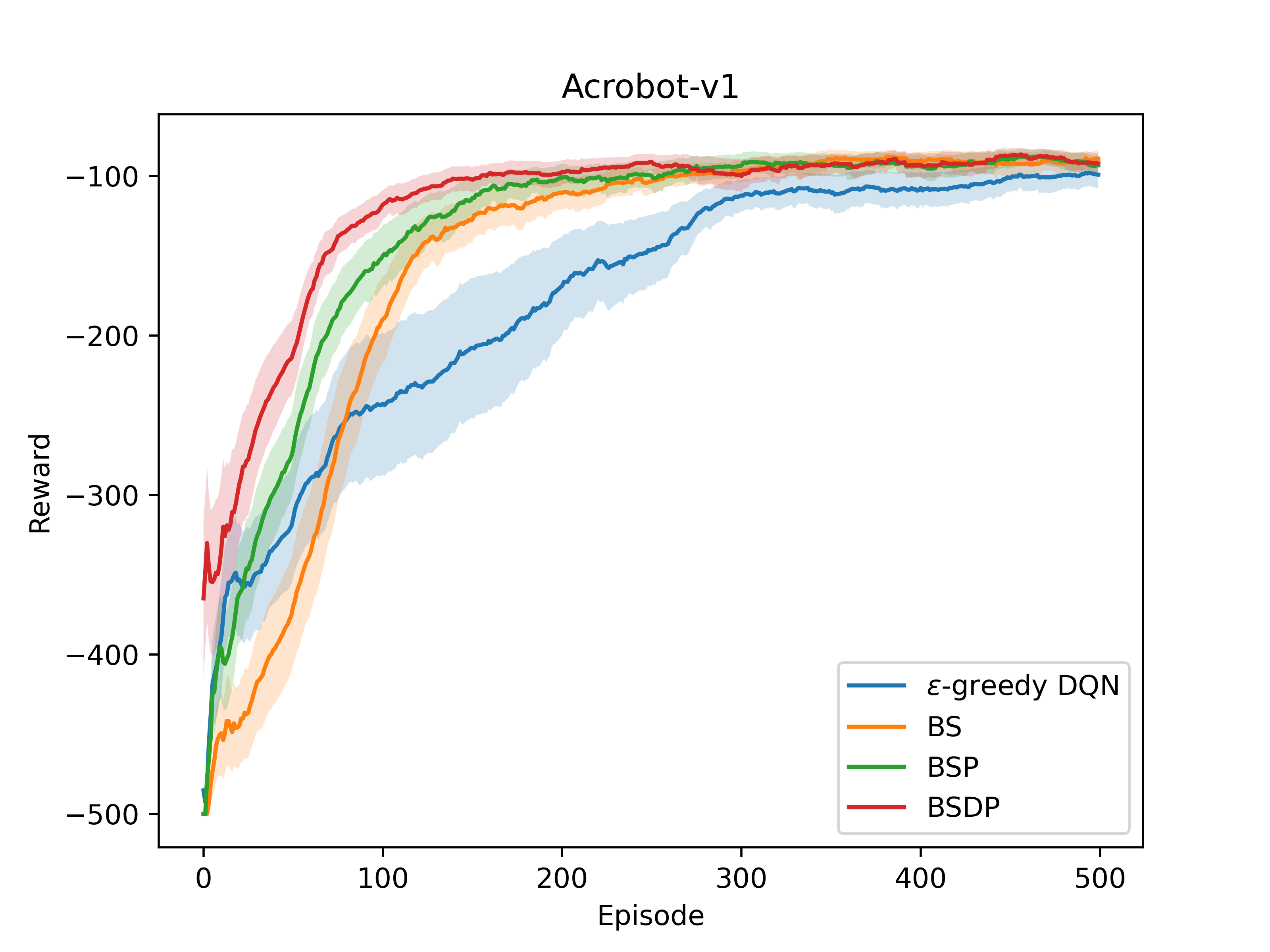}}
    \hfill
  \subfloat[Acrobot: E-rate \label{fig:res1:subfigB}]{%
        \includegraphics[width=0.5\linewidth]{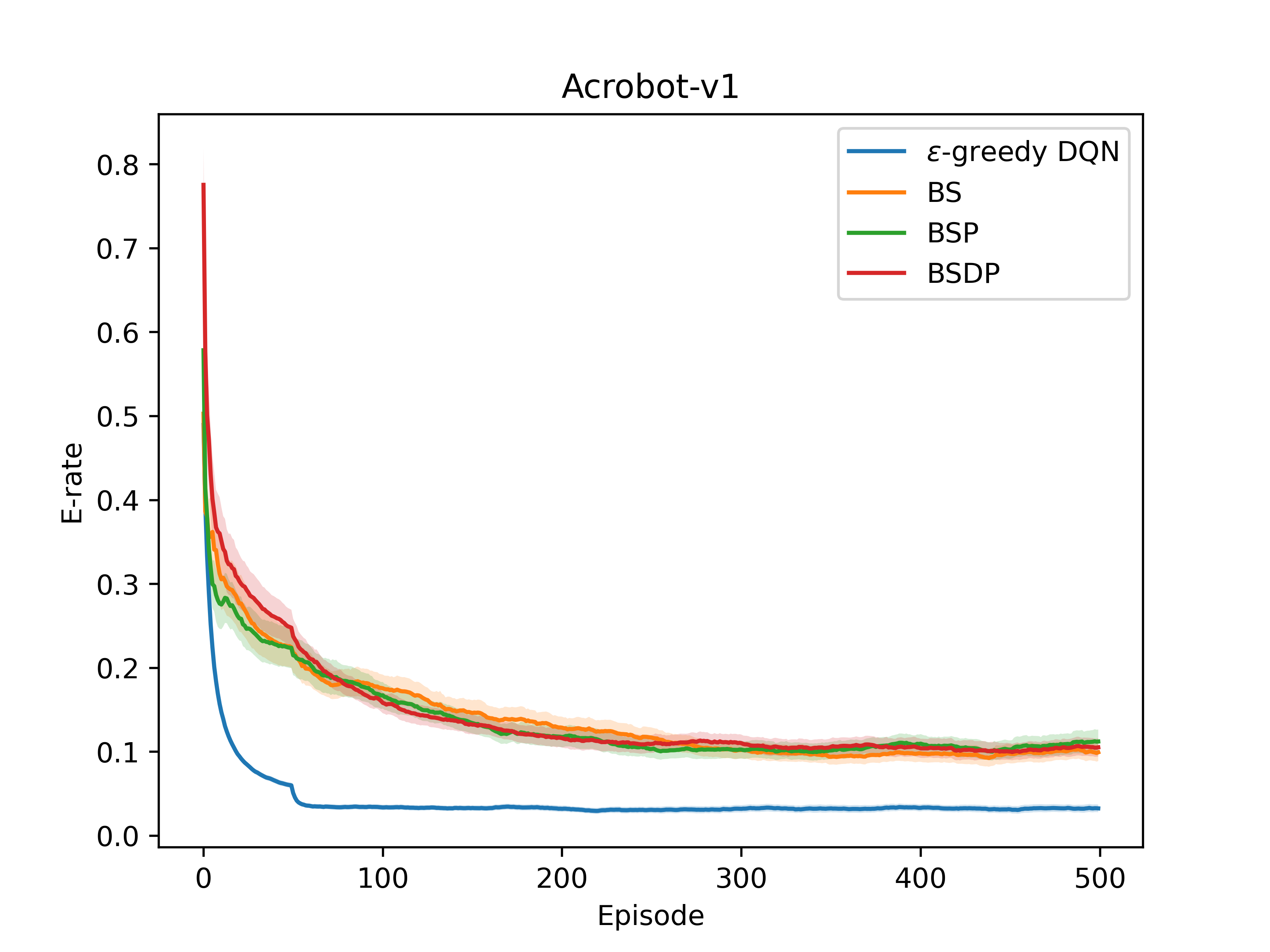}}
    \\
  \subfloat[MountainCar: reward\label{fig:res1:subfigC}]{%
        \includegraphics[width=0.5\linewidth]{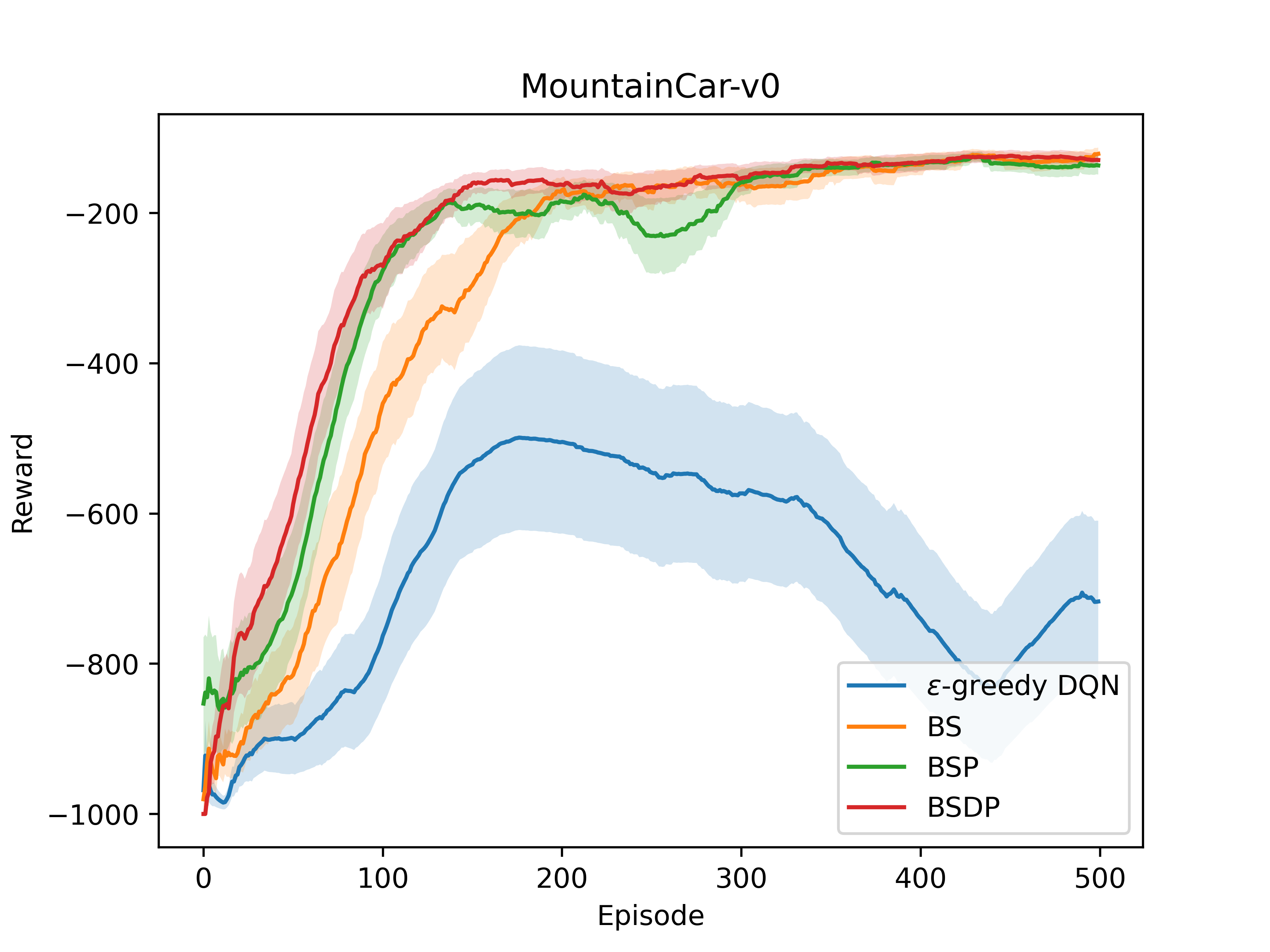}}
    \hfill
  \subfloat[MountainCar: E-rate\label{fig:res1:subfigD}]{%
        \includegraphics[width=0.5\linewidth]{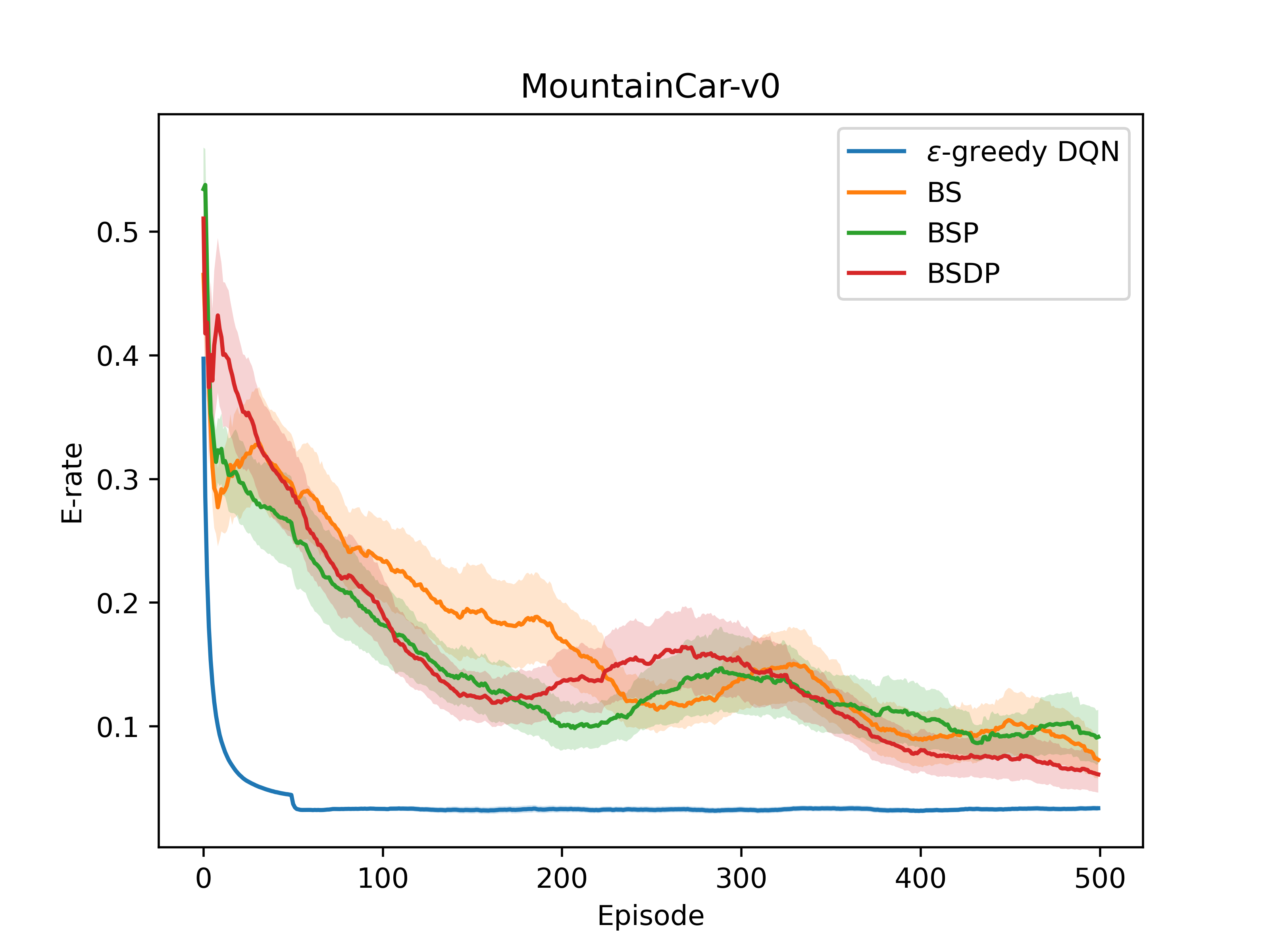}}
    \\
     \subfloat[CartPole: reward\label{fig:res1:subfigE}]{%
        \includegraphics[width=0.5\linewidth]{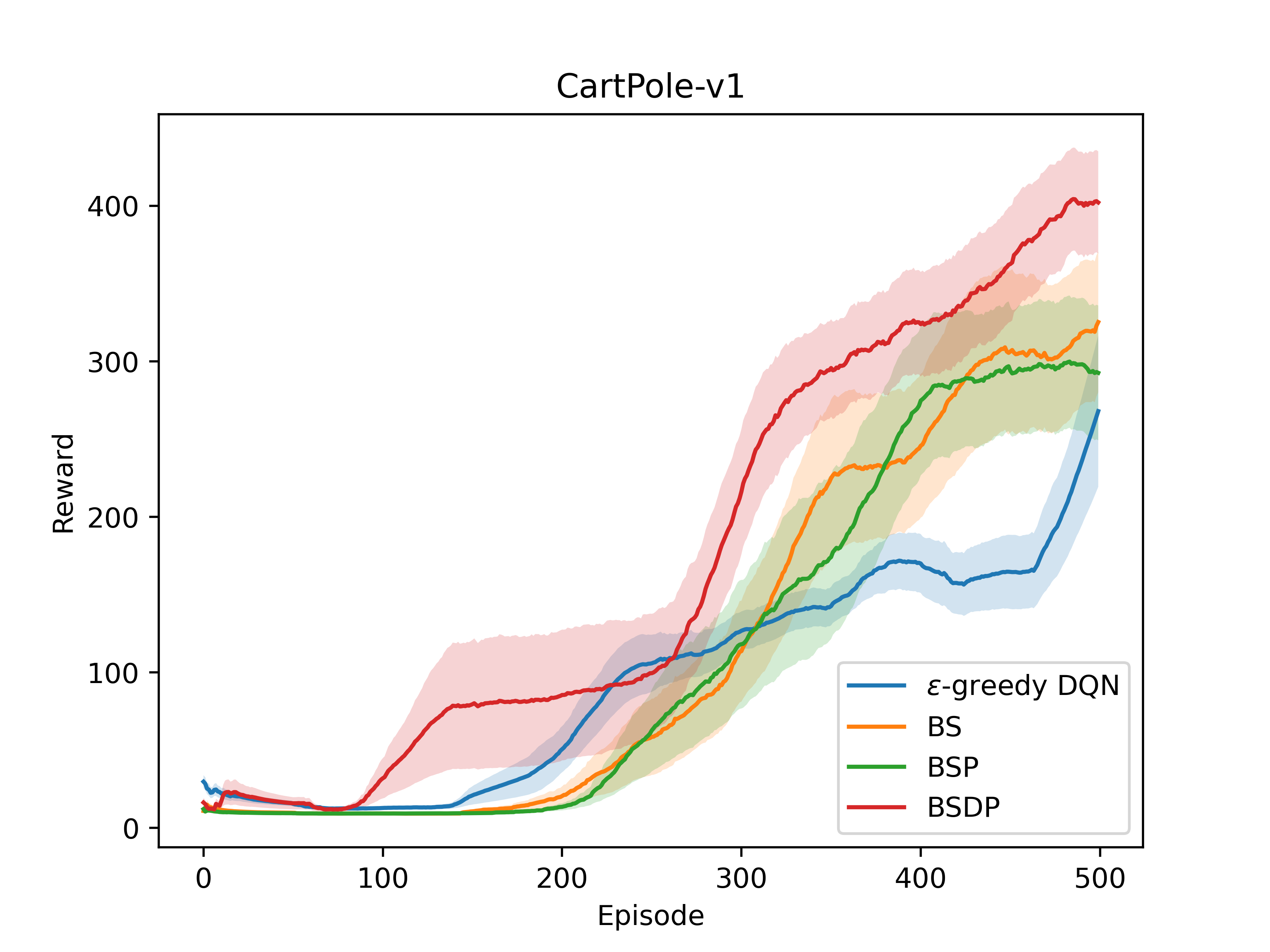}}
    \hfill
  \subfloat[CartPole: E-rate\label{fig:res1:subfigF}]{%
        \includegraphics[width=0.5\linewidth]{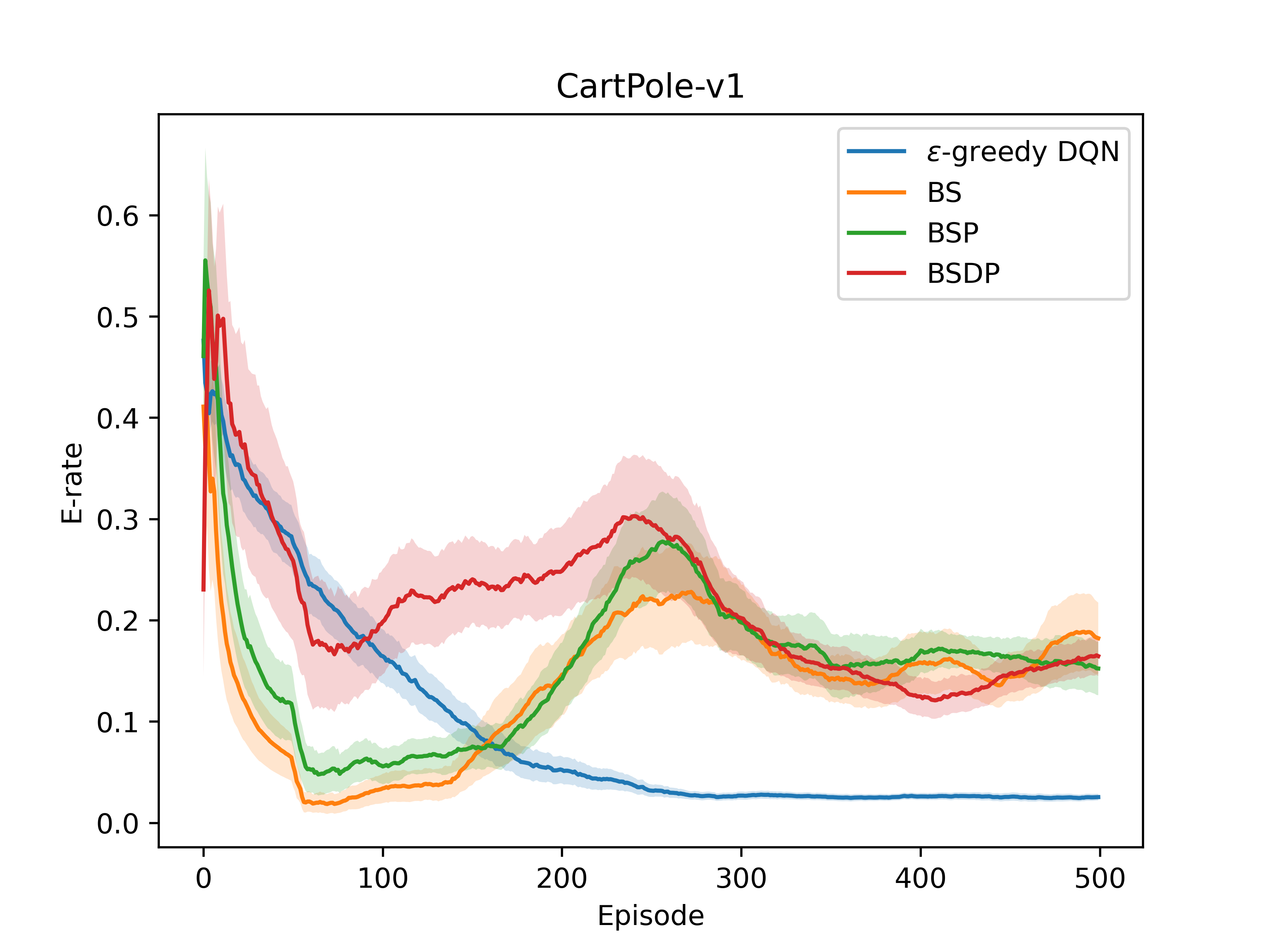}}
  \caption{Experiment results for investigating prior effect.}
  \label{fig:experiment1_result1}
\end{figure}

     Fig.~\ref{fig:res1:subfigA}, Fig.~\ref{fig:res1:subfigC} and Fig.~\ref{fig:res1:subfigE} show the reward curves over episodes for BS, BSP and BSDP in three \hyperref[classiccontrolenv]{classic control environments}, and Fig.~\ref{fig:res1:subfigB}, Fig.~\ref{fig:res1:subfigD} and Fig.~\ref{fig:res1:subfigF} depict the exploration rate\footnote{If the chosen action differs from the pure exploitation action, it is deemed an exploration action. The exploration rate is the ratio of exploration actions in an episode. For the ensemble-based method, the pure exploitation action corresponds to the action with the highest $Q$-value derived from the ensemble Q's average outputs.} curves over episodes. 
     
     In three classic control environments, it is observed that the ensemble-based methods (BS, BSP and BSDP) generally outperformed the $\epsilon$-greedy DQN. In these three ensemble-based methods, BSDP converged fastest and achieved high reward in a relatively smaller number of episodes. BSP showed improvement over BS at the beginning of training in Acrobot and MountainCar while achieving similar results to BS in CartPole. All these three converged to a similar reward level ($-100$ and $-130$) after extensive episodes of training in Acrobot (Fig.~\ref{fig:res1:subfigA}) and MountainCar (Fig.~\ref{fig:res1:subfigC}). Although the results in Cartpole (Fig.~\ref{fig:res1:subfigE}) did not demonstrate similar episodic rewards at the end of the training, we believe that the three curves will converge with the proceeding of training as they share the same underlying mechanism. This aligns with Bayesian principles as the impact of prior functions will diminish with the incorporation of new information.

    The exploration rate is closely related to the exploration behavior of the agent and its performance of episodic cumulative reward. For decaying $\epsilon$-greedy exploration strategy, the exploration rate decreases deterministically as the total steps increase. In contrast, all the ensemble-based methods that use Thompson Sampling as an exploration strategy exhibit an adaptive exploration rate. For example, in CartPole, BSDP's exploration rates first decreased to $0.17$, then increased to $0.3$ and finally converged to a low value of $0.15$. This exploration behavior is determined by the uncertainty among all the ensemble members. If most members show a high $Q$-value on the same action, then the agent will tend to exploit rather than explore. On the other hand, if members have a large disagreement in the Q-value of a state/action, the agent will tend to explore to reduce knowledge uncertainty among the members. 
    
    Among the three ensemble-based methods, BSDP demonstrates the highest exploration rate at the beginning of the training due to the diversity introduced in the priors, which makes the initial uncertainty high. This high uncertainty at the beginning of the training has improved the sample efficiency as the reward curves demonstrated. It is because incorporating diversity priors has mitigated the overconfidence of decision-making in the early stage, which consequently leads to more sufficient exploration. Considering MountainCar as an example: before reaching the goal, agents encounter only zero rewards. This causes all the $Q$-values in the ensemble to be updated towards zero, leading to a rapid decrease in uncertainty. However, this reduction in uncertainty is misleading, as agents are still uncertain about the true goal or the presence of the positive rewards in the environment. Diversity initialization can counteract this rapid reduction in uncertainty by enhancing the initial dissimilarity among ensemble members. The BSDP's E-rate curve in MountainCar (Fig.~\ref{fig:res1:subfigD}) also supported the underlying rationale, which immediately slowed the quick reduction of uncertainty and maintained a high E-rate ($0.4$) in the first 30 episodes while BSP just maintained around $0.3$ E-rate. 
    
\section{Conclusion}
Our research has significant implications for uncertainty-based exploration in RL, especially for those using the ensemble method to estimate uncertainty. In this paper, we propose a bootstrapped DQN with diverse prior for active exploration, which can effectively address the challenge of uncertainty estimation bias at the beginning of the RL training. BSDP increases the initial diversity among ensemble members, which mitigates the underestimation of uncertainty in the early stage of training and also serves as an intrinsic motivation to drive the agent's initial exploration. Consequently, BSDP achieves superior sample efficiency and higher early rewards compared to both no prior (BS) and random prior (BSP) methods in classic control and BinaryChain problems. 

\bibliographystyle{IEEEtran}
\bibliography{main}{}

\begin{thebibliography}{10}
\providecommand{\url}[1]{#1}
\csname url@samestyle\endcsname
\providecommand{\newblock}{\relax}
\providecommand{\bibinfo}[2]{#2}
\providecommand{\BIBentrySTDinterwordspacing}{\spaceskip=0pt\relax}
\providecommand{\BIBentryALTinterwordstretchfactor}{4}
\providecommand{\BIBentryALTinterwordspacing}{\spaceskip=\fontdimen2\font plus
\BIBentryALTinterwordstretchfactor\fontdimen3\font minus \fontdimen4\font\relax}
\providecommand{\BIBforeignlanguage}[2]{{%
\expandafter\ifx\csname l@#1\endcsname\relax
\typeout{** WARNING: IEEEtran.bst: No hyphenation pattern has been}%
\typeout{** loaded for the language `#1'. Using the pattern for}%
\typeout{** the default language instead.}%
\else
\language=\csname l@#1\endcsname
\fi
#2}}
\providecommand{\BIBdecl}{\relax}
\BIBdecl

\bibitem{silver2016mastering}
D.~Silver, A.~Huang, C.~J. Maddison, A.~Guez, L.~Sifre, G.~Van Den~Driessche, J.~Schrittwieser, I.~Antonoglou, V.~Panneershelvam, M.~Lanctot \emph{et~al.}, ``Mastering the game of go with deep neural networks and tree search,'' \emph{Nature}, vol. 529, no. 7587, pp. 484--489, 2016.

\bibitem{pmlr-v119-badia20a}
A.~P. Badia, B.~Piot, S.~Kapturowski, P.~Sprechmann, A.~Vitvitskyi, Z.~D. Guo, and C.~Blundell, ``Agent57: Outperforming the {A}tari human benchmark,'' in \emph{Proceedings of the 37th International Conference on Machine Learning}, ser. Proceedings of Machine Learning Research, vol. 119.\hskip 1em plus 0.5em minus 0.4em\relax PMLR, 13--18 Jul 2020, pp. 507--517.

\bibitem{lillicrap2015continuous}
T.~P. Lillicrap, J.~J. Hunt, A.~Pritzel, N.~Heess, T.~Erez, Y.~Tassa, D.~Silver, and D.~Wierstra, ``Continuous control with deep reinforcement learning,'' \emph{arXiv preprint arXiv:1509.02971}, 2015.

\bibitem{cox1979theoretical}
D.~R. Cox and D.~V. Hinkley, \emph{Theoretical Statistics}.\hskip 1em plus 0.5em minus 0.4em\relax CRC Press, 1979.

\bibitem{wald1950statistical}
A.~Wald, ``Statistical decision functions.'' 1950.

\bibitem{jospin2022hands}
L.~V. Jospin, H.~Laga, F.~Boussaid, W.~Buntine, and M.~Bennamoun, ``Hands-on bayesian neural networks—a tutorial for deep learning users,'' \emph{IEEE Computational Intelligence Magazine}, vol.~17, no.~2, pp. 29--48, 2022.

\bibitem{lakshminarayanan2017simple}
B.~Lakshminarayanan, A.~Pritzel, and C.~Blundell, ``Simple and scalable predictive uncertainty estimation using deep ensembles,'' \emph{Advances in Neural Information Processing Systems}, vol.~30, 2017.

\bibitem{osband2016deep}
I.~Osband, C.~Blundell, A.~Pritzel, and B.~Van~Roy, ``Deep exploration via bootstrapped {DQN},'' \emph{Advances in Neural Information Processing Systems}, vol.~29, 2016.

\bibitem{osband2018randomized}
I.~Osband, J.~Aslanides, and A.~Cassirer, ``Randomized prior functions for deep reinforcement learning,'' \emph{Advances in Neural Information Processing Systems}, vol.~31, 2018.

\bibitem{fortuin2022priors}
V.~Fortuin, ``Priors in bayesian deep learning: A review,'' \emph{International Statistical Review}, vol.~90, no.~3, pp. 563--591, 2022.

\bibitem{chentanez2004intrinsically}
N.~Chentanez, A.~Barto, and S.~Singh, ``Intrinsically motivated reinforcement learning,'' \emph{Advances in Neural Information Processing Systems}, vol.~17, 2004.

\bibitem{bellemare2016unifying}
M.~Bellemare, S.~Srinivasan, G.~Ostrovski, T.~Schaul, D.~Saxton, and R.~Munos, ``Unifying count-based exploration and intrinsic motivation,'' \emph{Advances in Neural Information Processing Systems}, vol.~29, 2016.

\bibitem{bellemare2013arcade}
M.~G. Bellemare, Y.~Naddaf, J.~Veness, and M.~Bowling, ``The arcade learning environment: An evaluation platform for general agents,'' \emph{Journal of Artificial Intelligence Research}, vol.~47, pp. 253--279, 2013.

\bibitem{ghavamzadeh2015bayesian}
M.~Ghavamzadeh, S.~Mannor, J.~Pineau, A.~Tamar \emph{et~al.}, ``Bayesian reinforcement learning: A survey,'' \emph{Foundations and Trends{\textregistered} in Machine Learning}, vol.~8, no. 5-6, pp. 359--483, 2015.

\bibitem{ladosz2022exploration}
P.~Ladosz, L.~Weng, M.~Kim, and H.~Oh, ``Exploration in deep reinforcement learning: A survey,'' \emph{Information Fusion}, vol.~85, pp. 1--22, 2022.

\bibitem{cowen2020samba}
A.~I. Cowen-Rivers, D.~Palenicek, V.~Moens, M.~Abdullah, A.~Sootla, J.~Wang, and H.~Ammar, ``Samba: Safe model-based \& active reinforcement learning,'' \emph{arXiv preprint arXiv:2006.09436}, 2020.

\bibitem{lindner2021information}
D.~Lindner, M.~Turchetta, S.~Tschiatschek, K.~Ciosek, and A.~Krause, ``Information directed reward learning for reinforcement learning,'' \emph{Advances in Neural Information Processing Systems}, vol.~34, pp. 3850--3862, 2021.

\bibitem{gawlikowski2023survey}
J.~Gawlikowski, C.~R.~N. Tassi, M.~Ali, J.~Lee, M.~Humt, J.~Feng, A.~Kruspe, R.~Triebel, P.~Jung, R.~Roscher \emph{et~al.}, ``A survey of uncertainty in deep neural networks,'' \emph{Artificial Intelligence Review}, pp. 1--77, 2023.

\bibitem{renda2019comparing}
A.~Renda, M.~Barsacchi, A.~Bechini, and F.~Marcelloni, ``Comparing ensemble strategies for deep learning: An application to facial expression recognition,'' \emph{Expert Systems with Applications}, vol. 136, pp. 1--11, 2019.

\bibitem{livieris2021ensemble}
I.~E. Livieris, L.~Iliadis, and P.~Pintelas, ``On ensemble techniques of weight-constrained neural networks,'' \emph{Evolving Systems}, vol.~12, pp. 155--167, 2021.

\bibitem{nalepa2019training}
J.~Nalepa, M.~Myller, and M.~Kawulok, ``Training-and test-time data augmentation for hyperspectral image segmentation,'' \emph{IEEE Geoscience and Remote Sensing Letters}, vol.~17, no.~2, pp. 292--296, 2019.

\bibitem{herron2020ensembles}
E.~J. Herron, S.~R. Young, and T.~E. Potok, ``Ensembles of networks produced from neural architecture search,'' in \emph{International Conference on High Performance Computing}.\hskip 1em plus 0.5em minus 0.4em\relax Springer, 2020, pp. 223--234.

\bibitem{peer2021ensemble}
O.~Peer, C.~Tessler, N.~Merlis, and R.~Meir, ``Ensemble bootstrapping for q-learning,'' in \emph{International Conference on Machine Learning}.\hskip 1em plus 0.5em minus 0.4em\relax PMLR, 2021, pp. 8454--8463.

\bibitem{van2016deep}
H.~Van~Hasselt, A.~Guez, and D.~Silver, ``Deep reinforcement learning with double q-learning,'' in \emph{Proceedings of the AAAI Conference on Artificial Intelligence}, vol.~30, no.~1, 2016.

\bibitem{he2015delving}
K.~He, X.~Zhang, S.~Ren, and J.~Sun, ``Delving deep into rectifiers: Surpassing human-level performance on imagenet classification,'' in \emph{Proceedings of the IEEE international Conference on Computer Vision}, 2015, pp. 1026--1034.

\bibitem{pathak2019self}
D.~Pathak, D.~Gandhi, and A.~Gupta, ``Self-supervised exploration via disagreement,'' in \emph{International Conference on Machine Learning}.\hskip 1em plus 0.5em minus 0.4em\relax PMLR, 2019, pp. 5062--5071.

\bibitem{kaelbling1996reinforcement}
L.~P. Kaelbling, M.~L. Littman, and A.~W. Moore, ``Reinforcement learning: A survey,'' \emph{Journal of Artificial Intelligence research}, vol.~4, pp. 237--285, 1996.

\bibitem{russo2018tutorial}
D.~J. Russo, B.~Van~Roy, A.~Kazerouni, I.~Osband, Z.~Wen \emph{et~al.}, ``A tutorial on thompson sampling,'' \emph{Foundations and Trends{\textregistered} in Machine Learning}, vol.~11, no.~1, pp. 1--96, 2018.

\bibitem{efron1982jackknife}
B.~Efron, \emph{The jackknife, the bootstrap and other resampling plans}.\hskip 1em plus 0.5em minus 0.4em\relax SIAM, 1982.

\bibitem{gymlibrary}
``Gym documentation,'' \url{https://www.gymlibrary.dev/}, 2022, accessed: October 14, 2023.

\end{thebibliography}

\end{document}